\documentclass[journal]{IEEEtran}
\usepackage{cite}

\usepackage{amsmath}
\usepackage{url}
\usepackage{hyperref}
\usepackage[pdftex]{graphicx}
\usepackage{subfigure}
\usepackage{multirow}
\usepackage{adjustbox}
\usepackage{float}
\usepackage{placeins}

\usepackage{xcolor}
\definecolor{fuchsiapink}{rgb}{1.0, 0.47, 1.0}
\definecolor{tractorred}{rgb}{0.99, 0.05, 0.21}
\definecolor{forestgreen}{rgb}{0.13, 0.55, 0.13}
\definecolor{pink}{rgb}{0.9, 0.2, 0.7}

\newcommand{\changed}[1]{\textcolor{black}{#1}}

\newcommand{\latentmethod}{LatentIM}
\newcommand{\outcomemethod}{OutcomeIM}
\newcommand{\randommethod}{RandomIM}
\newcommand{\objectmethod}{ObjectIM}
\newcommand{\actionmethod}{ActionIM}
\newcommand{\appref}[1]{\hyperref[#1]{Appendix}}

\begin{document}
%

\title{\changed{Exploration with Intrinsic Motivation using Object-Action-Outcome Latent Space}}
%
%
%
%

\author{\IEEEauthorblockN{Melisa Idil Sener$^{1}$, Yukie Nagai$^{2}$, Erhan Oztop$^{3,4}$ and Emre Ugur$^{1}$}
				\vspace{0.5cm}
				
				\IEEEauthorblockA{ $^1$Bogazici University, Istanbul, Turkey.}
				\IEEEauthorblockA{ $^2$The University of Tokyo, Tokyo, Japan.} 
				\IEEEauthorblockA {$^3$Osaka University, Osaka, Japan.}
				\IEEEauthorblockA {$^4$Ozyegin University, Istanbul, Turkey.}
}

%



\maketitle

\begin{abstract}

One effective approach for equipping artificial agents with sensorimotor skills is to use self-exploration. To do this efficiently is critical, as time and data collection are costly. In this study, we propose an exploration mechanism that blends action, object, and action outcome representations into a latent space, where local regions are formed to host forward model learning. The agent uses intrinsic motivation to select the forward model with the highest learning progress to adopt at a given exploration step. This parallels how infants learn, as high learning progress indicates that the learning problem is neither too easy nor too difficult in the selected region. The proposed approach is validated with a simulated robot in a table-top environment. \changed{The simulation scene comprises a robot and various objects, where the robot interacts with one of them each time using a set of parameterized actions and learns the outcomes of these interactions.} With the proposed approach, the robot organizes its curriculum of learning as in existing intrinsic motivation approaches and outperforms them in learning speed. Moreover, the learning regime demonstrates features that partially match infant development; in particular, the proposed system learns to predict \changed{the outcomes of different skills in a staged manner.}
\end{abstract}


%
\IEEEpeerreviewmaketitle

\section{Introduction}
\IEEEPARstart{F}{rom} the moment they are born, babies begin learning about their bodies and the environment autonomously. Even when there is no immediate reward or explicit assistance from their caregiver, it is quite interesting that they conduct this learning process and develop sophisticated skills. Autonomous exploration has been regarded as an essential mechanism for the learning and development of living organisms \cite{oudeyer2009intrinsic, cangelosi2015developmental}. Exploratory behaviors, which enable us to adapt to different kinds of situations, learn complex skills, and practice our creativity, are observed not only in humans but also in other animals \cite{white1959motivation, berlyne1966curiosity}. \changed{Because of the need to feel competent and self-determining, humans engage in novelty-seeking behaviors such as exploration and play \cite{deci_intrinsic_1975}, which are later described with the generic term of ``intrinsically motivated" behavior\cite{deci1997m}.} \changed{Recently, the neural correlates of such behaviors have been found to be linked to dopaminergic systems in the brain (see \cite{GottliebLopesOudeyer2016,oudeyer2016intrinsic}).} 

Intrinsically motivated strategies have been used along with various types of robot learning methods and applications such as socially guided learning \cite{ivaldi2013learning, duminy2019learning, fournier2019clic}, affordances \cite{ugur2016emergent, manoury2019hierarchical, baldassarre2019embodied}, and planning \cite{blaes2019control}. Given the exploration space of the agent, a particular intrinsic motivation (IM) signal, \textit{learning progress} \cite{oudeyer2007intrinsic, schmidhuber1991curious}, aims to give priority to exploration regions that are neither too easy nor too difficult to learn, i.e., with the appropriate level of complexity which is inline with infant data\cite{kidd2012goldilocks}.

Inspired by infant development, this paper studies how a manipulator robot can learn the outcomes of its actions via autonomous exploration and intrinsic motivation. Predicting the consequences of own actions is an important requirement for intelligent control and decision making in both biological and artificial systems. The importance of predictive learning in human sensorimotor and cognitive development has already been emphasized by \cite{nagai2019predictive}. The exploration space of the manipulator robot for predictive learning is composed of the space of objects that it encounters, the action space of the robot, and the outcomes that it observes. During its exploration, the robot is expected to select objects, actions, and outcomes intelligently in order to acquire the target prediction capability most efficiently. It is desirable to avoid pre-defining the set of objects, actions, and outcomes in unsupervised learning settings, where they are typically represented or parameterized by continuous variables. Therefore, the robot has to explore a continuous space \changed{with the help of IM} to form predictive \changed{internal} models that can be used for better control and decision-making. How animals and humans \changed{efficiently and effectively construct predictive internal models for representing high-dimensional object-action-outcome relations} inspires our approach as well as several other computational approaches \cite{jacobs1991adaptive, wolpert1998multiple, nagai2019predictive}. \changed{It is generally accepted that humans and animals are endowed with neural mechanisms that build internal models to facilitate effective control of objects and/or body parts in different dynamics \cite{kawato1999internal}.}  An internal model is a computational structure that mimics (a part of) the sensorimotor system in terms of input-output relations, which may be conceived at \changed{different levels of motor control hierarchy (see, e.g., \cite{Cui2016forwardmodel_ppc,oztopWolpertKawato2005}).} For motor control, Wolpert and Kawato \cite{wolpert1998multiple} proposed a computational model that is composed of multiple paired forward-inverse models. The contribution of each pair to the behavioral output is determined by a \textit{responsibility signal} that is computed based on the model's prediction ability. The general benefits for adopting such a modular strategy are suggested as (1) \changed{efficient coding of the tasks that might be encountered in a variety of contexts; (2) simultaneous learning of such task} contexts without interference; and (3) the possibility of learning a more complex context by reusing the knowledge captured in the learned modules.

In this study, we also adopt a modular approach for forward modeling, and use learning progress measure to gate learning. To form the modules, the exploration space of the robot, i.e., object-action-outcome space, is transformed into a compact latent space and then partitioned into regions, for which individual forward models are trained to become responsible for their region. Directly partitioning the object-action-outcome space is not feasible through standard clustering algorithms as this space is composed of diverse and complex set of variables such as pixel values of the top-down depth image of the objects, various parameters of the manipulation actions, and the position and orientation changes. For effectively partitioning the exploration space, first, a low-dimensional latent representation, that fuses the related triplets (object, action, outcome), is formed. Then, the formed  latent space is clustered into regions. During forward model learning, the regions with the highest learning progress, i.e., the regions whose forward models exhibit a maximum decline in prediction error, are prioritized. Through simulation experiments  involving a robot arm-hand system that reaches and grasps different types of objects placed in various orientations and sizes with varying arm and hand parameters, we showed that: 
\begin{itemize}
    \item the exploration regions formed in the blended object-action-outcome space correspond to semantically meaningful manipulation primitives,
    \item the proposed latent space IM approach outperforms competing IM methods \changed{(adapted to our setup)} that utilize only object\cite{ugur2007curiosity}, action\cite{ugur2016emergent}, or outcome/goal\cite{baranes2013active} spaces in terms of learning speed, and
    \item \changed{the exploration order that the proposed approach posits is partially parallel with the staged development} of action prediction in infants. To be concrete, the proposed system learns to predict the basic grasp action outcomes before learning the outcome of purposeful push actions \cite{scharf2016developmental}.
\end{itemize}

The rest of this paper is structured as follows: the related literature is first reviewed in Section \ref{sec:relatedwork}. Then, the proposed architecture, including its components and the experimental setup, are presented in Sections \ref{sec:proposedsystem} and \ref{sec:experimentsetup}. Section \ref{sec:results} demonstrates the outperforming results of the proposed system. Finally, Section \ref{sec:discussion} gives discussion and conclusions. 

\section{Related Work}\label{sec:relatedwork}

\subsection{Computational Models of Intrinsic Motivation}

Regarding the high-dimensional and complex dynamics involved in physical systems, exploration is considered an essential problem in robot learning \cite{lopes2010guest}. \changed{Intrinsically motivated strategies are widely used to address the exploration problem. Oudeyer and Kaplan \cite{oudeyer2009intrinsic} divide the computational approaches of IM into two classes as \textit{Knowledge-Based IM (KB-IM)} and \textit{Competence-Based IM (CB-IM)}. The KB-IM strategy is derived from the deviation of the agent's knowledge of the environment from reality \cite{oudeyer2009intrinsic}. The agent learns new skills while expanding its knowledge about the environment by exploring the situations outside of its current understanding \cite{santucci2013best}. The CB-IM strategy focuses on a specific state, i.e., the goal state, that changes adaptively according to the current competencies of the agent \cite{santucci2013best}. Mirolli and Baldassarre \cite{mirolli2013functions} state that both KB-IM and CB-IM can serve knowledge and competence acquisition. In other words, KB and CB-IM have a complementary relationship, where the former aims to detect which skills to train based on their novelties, and the latter to select the expert to achieve a particular goal, as shown in \cite{rayyes2020efficient}. In terms of robotic application, intrinsic motivation has been studied under reinforcement learning \cite{chentanez2005intrinsically, mohamed2015variational} and developmental robotics \cite{blank2005bringing, oudeyer2007intrinsic, baranes2009r, moulin2014explauto, forestier2016modular}.}

\paragraph{Reinforcement Learning (RL)} In reinforcement learning, an agent learns an optimal policy to accomplish certain goals typically by considering the extrinsic rewards, i.e., external rewards that stem from the task definition. However, in some settings, the extrinsic reward may be absent or sparse. Even if that is the case, an autonomous agent should be able to learn skills. To deal with this situation, intrinsic rewards are used in several RL studies. Some studies used only intrinsic reward \cite{mohamed2015variational, hester2017intrinsically, eysenbach2018diversity}, whereas others studied how to combine intrinsic and extrinsic rewards in RL settings \cite{chentanez2005intrinsically, hafez2019deep, blau2019bayesian}. Intrinsic motivation was also applied to RL at the different levels of the hierarchies \cite{kulkarni2016hierarchical, vezhnevets2017feudal, nachum2018data}. All of these studies aim to make the agent learn skills to achieve a specific goal. By contrast, in our study, there is no particular goal that the agent needs to accomplish.  

\paragraph{Developmental Robotics} In the seminal computational architecture of Oudeyer et al.\cite{oudeyer2007intrinsic}, the sen\-so\-rimotor space was incrementally split into regions, and the regions were learned by the local experts. Selection between the regions was made by considering the learning progress IM signal. In our previous work \cite{sener2018partitioning}, similar to \cite{oudeyer2007intrinsic}, we partitioned the sensorimotor space by considering a single parameter at each partitioning step in order to form exploration regions. Forestier et al.\cite{forestier2017intrinsically} developed an algorithmic procedure called ``intrinsically motivated goal exploration processes'' (IMGEP) that allow\changed{s} the autonomous discovery of skills in an open-ended learning setting. In their approach, the agent selected the goal to pursue using intrinsic motivation signal and learned skills by self-experimentation. As a result, the agent learned to discover and accomplish goals by following a self-generated curriculum with an increasing level of complexity. Mannella et al.\cite{mannella2018know} hypothesized that an agent learns the dynamics of its body by autonomous goal generation regulated by intrinsic motivation. To validate their hypothesis, they created a model that relied on an intrinsic motivation signal to form abstract representations of the observations and select goals to pursue and learn motor skills. Haber et al.\cite{haber2018emergence} proposed a computational model of intrinsic motivation where the understanding of ego-motion, followed by the ability to interact with single and multiple objects, emerges from novelty-seeking exploration.  In our current work, different from the previous studies, IM-exploration regions are formed by clustering a latent space that combines object, action, and outcome information.

\subsection{Representation Learning in Robotics and IM}
Most of the work in robot learning utilizes engineered feature representations to perform given tasks. However, to obtain full autonomy in intelligent systems, the agent also should be capable of building efficient feature representations from raw sensory data. Representation learning in robotics is an important research direction that allows the learning systems to be efficient in computational resources, generalization ability, time efficiency, and abandons the need for feature engineering. Various studies in domains of robot learning \cite{jonschkowski2014state, jonschkowski2015learning}, planning \cite{boots2011closing, banijamali2018robust}, control \cite{lange2012autonomous, watter2015embed, banijamali2018robust}, and RL \cite{mohamed2015variational, cruz2016multi, santucci2016grail, vezhnevets2017feudal, hafez2019deep, hafez2019efficient, blau2019bayesian, Whitney2020Dynamics-Aware} focus on learning representations to foster autonomy. Among these, a number of studies utilized representation learning in IM-based exploration \cite{mohamed2015variational, santucci2016grail, laversanne2018curiosity, pere2018unsupervised, manoury2019chime, schillaci2020intrinsic}. Bugur et al. \cite{bugur2019effect} proposed an intrinsically motivated exploration scheme in action space. In their study, the action and effect space information was used to obtain a latent representation from which two regions are obtained for exploration via IM. Laversanne-Finot et al.\cite{laversanne2018curiosity} integrated a representation learning stage on top of IMGEP \cite{forestier2017intrinsically} to create the goal-spaces by encoding raw sensory observations. In that study, the agent first passively observes the environment to collect data for learning an embedding function. After that stage, learned representation was used to form goal spaces to be explored by the intrinsically motivated architecture they proposed previously. Hafez et al.\cite{hafez2019deep} proposed an Actor-Critic algorithm that enables the learning of motor skills directly from visual observations in an RL setting. In their work, an embedding of visual input was used in actor and critic networks to create exploration regions incrementally utilizing Self-Organizing Maps. Like our work, each region has a prediction model whose learning progress is then used to guide the exploration. In summary, almost all these studies considered only the observation space to form the latent space, and \cite{bugur2019effect} considered only action and effect space. In contrast, we exploit a latent representation that integrates high-dimensional object features, action parameters, and outcome observations in region formation and IM-based exploration. 

\begin{figure*}[ht!]
    \centering
    \includegraphics[width=\textwidth]{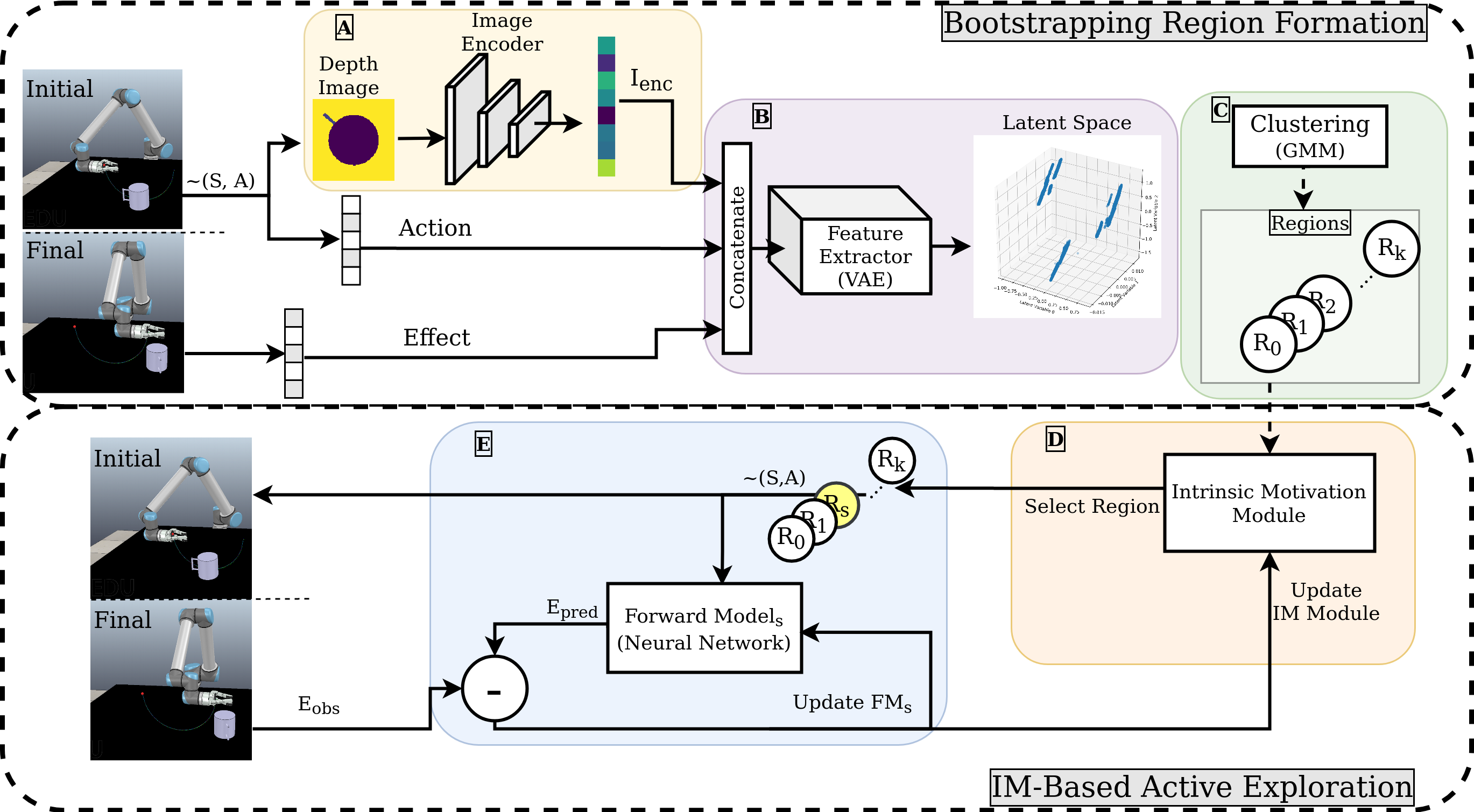}
    \caption{The overview of the proposed framework and learning cycle. The regions are formed via random exploration, as shown in the upper panel, and actively selected for exploration by the IM module, as shown in the lower panel. See the details in the text.}
    \label{fig:m_general}
\end{figure*}{}

\section{Proposed system}\label{sec:proposedsystem}
\subsection{Overview and General Flow}\label{ssec:m_general}

Fig.~\ref{fig:m_general} illustrates the general framework and the learning cycle proposed in this study. Recall that our aim is to partition the object-action-outcome exploration space of the robot into regions and enable the robot to explore these regions in the most efficient way via IM. The upper panel of the figure shows how these regions are formed in a bootstrapping phase, and the lower panel shows how these regions are selected in each IM-based exploration step. As shown in the upper panel, to bootstrap the region formation, the simulated robot (shown on the left) undergoes a short exploration phase, in which it interacts with a set of objects via randomly parameterized actions and observes the outcome of its actions. In each interaction, the information of the object  (depth image), action  (arm and hand parameters), and outcome  (change in object position and orientation) are collected. 
Using the data set obtained from these exploratory random interactions, the regions for predictive learning are found in two steps\changed{.} First, the processed depth image (shown in (A)), action, and outcome features are blended together and mapped to a low-dimensional latent space, as shown in (B). Second, a clustering algorithm is applied to find regions for predictive learning in the latent space, as shown in (C). In the IM-based active learning phase, shown in the lower panel, a forward model that predicts the outcome given object and action features is trained for each region (E), and the region whose forward model exhibits the highest learning progress is selected for further learning (D). After a pair of object and action (parameter vector) is sampled from the selected region, the robot observes the outcome of the application of the sampled action (bottom-left) and updates the corresponding forward model (E) and the learning progress statistics of the region (D). 

\subsection{Object-Action-Outcome Representations \label{ssec:m_implement}}
In each interaction, the robot executes its parametric action on an object and observes the outcome. 
\begin{itemize}
    \item {\bf Object:} The top-down depth image of the object, taken before the execution of the action, is encoded through a Convolutional Autoencoder (CAE) into a low-dimensional feature vector (Fig.~\ref{fig:m_general}(A)), $I_{enc}$ (8D). Hence, the object information to the system is represented by this low dimensional feature vector.
    \item {\bf Action:} We assume that the robot is equipped with a basic movement capability involving the arm and the fingers, which we call the {\em reach and enclose} action. The action is parameterized and set to generate a semi-circular hand trajectory (see Fig.~\ref{fig:m_actionparam}), mimicking a human-like radial motion allowing basic object interactions. The robot action parameters vector (5D) controls the radius of the hand trajectory (1D), the direction of the approach towards the object (1D), and the end-effector state (3D). 
    \item {\bf Outcome:} The outcome of an action is defined as the position and orientation change of the object. Thus, it is represented by a vector (5D) composed of the position change in the three coordinate axes and (sin\&cos values of the) orientation change around the vertical axis. 
\end{itemize}

\subsection{Bootstrapping Region Formation}
{\bf Formation of the Latent Space}
The interaction experience obtained from a short random exploration phase is used to form the latent space. The object, action, and outcome vectors are concatenated in a single feature vector for each interaction and processed via a feature extractor to form the latent space that compactly represents these three elements of the interaction (Fig.~\ref{fig:m_general}(B)). As the feature extractor, a Variational Autoencoder (VAE) with Gaussian prior was used. Following the input layer (18D), the encoder part of the VAE has an intermediate layer (9D) with ReLU non-linearity, followed by a hidden layer (3D) that is split into $\mu(z)$ and $\sigma(z)$  so that the network output can be considered to represent a Gaussian distribution \cite{kingma2013auto}. The decoder part has a structure that is symmetrical to the encoder part. It takes $z$ that is sampled from the encoder's output and has an output layer with sigmoid non-linearity. Binary cross-entropy is used as the reconstruction loss, and the VAE loss is calculated as in \cite{kingma2013auto}. The VAE is trained with Adam \cite{kingma2014adam} optimizer with a batch size of 100. The latent space is formed by using the $\mu(z)$ from the encoder's output. 

{\bf Formation of the Exploration Regions}
To form the regions for forward model learning, the latent space is clustered using the Gaussian Mixture Model (GMM) algorithm with an empirically chosen number of clusters \changed{five in the current experiments, see \appref{apen} for an analysis on this parameter)}, (Fig.~\ref{fig:m_general}(C)) where each cluster corresponds to a ``region" $(R_i)$ that the robot can build a local forward model for action outcome prediction. Note that the regions found in this step were frozen and not changed during IM-based predictive learning for computational convenience.

\subsection{IM-based Active Exploration}

{\bf Local Prediction Models}
Each region $(R_i)$ found in the bootstrapping phase (Fig.~\ref{fig:m_general}(C)) is assigned to a forward model (FM) that is responsible for predicting the outcome given the object features ($I_{enc}$) and the action parameters in that region (Fig.~\ref{fig:m_general}(E)). The FMs are implemented as one hidden-layer feed-forward neural networks. Input, hidden, and output dimensions are set to 13, 512, and 5. The hidden unit non-linearity is provided by the ReLU activation function. At each predictive learning step, one FM is allowed to learn (see below). The learning in FMs is carried out by back-propagating the prediction error calculated as the mean square error (MSE). At each exploration step, a small batch $(\kappa)$ is sampled from the FM's responsibility region. In order to avoid \textit{catastrophic forgetting} \cite{mcclelland1995there, parisi2019continual}, the FM is continued training\footnote{FMs are implemented by using Keras \cite{chollet2015keras} deep learning library.} with all the data it has encountered so far. To avoid overfitting, the FM is trained for only a small number (5) of epochs at each step.

\begin{figure}[b!]
    \centering
    \includegraphics[width=\linewidth]{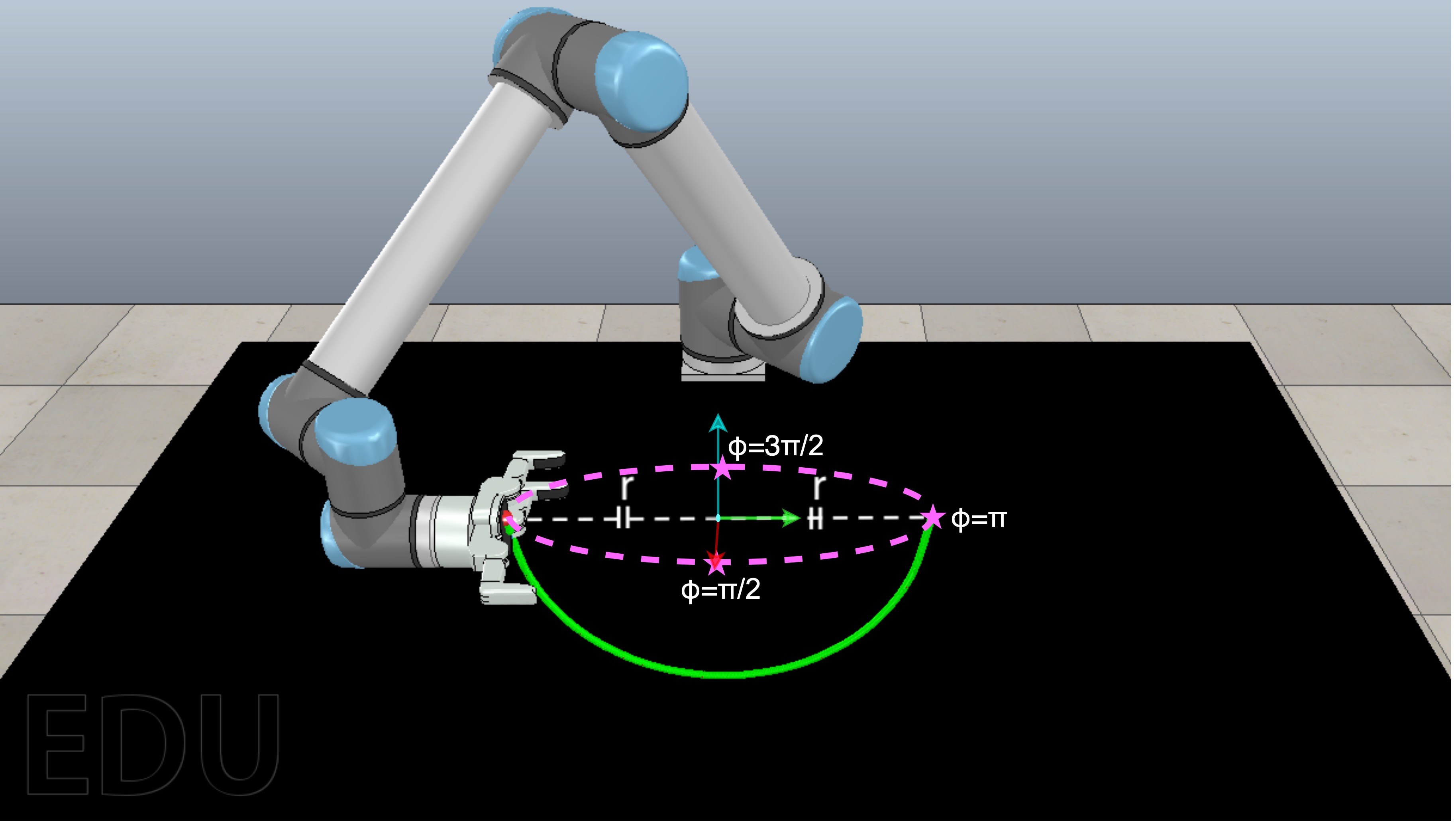}
    \caption{The end-effector follows a semi-circular trajectory above the surface of the table. The action parameters control the radius of the semi-circle $(r_{path})$ and the approach position of the end-effector to the object $(\phi_{path})$.}
    \label{fig:m_actionparam}
\end{figure}{}

{\bf Efficient Predictive Learning \label{ssec:m_IML}}
Learning progress based IM is used to select which region to target for improving the prediction ability  (Fig.~\ref{fig:m_general}(D)). Intrinsic Motivation Module keeps statistics about the (FM) learning progress of each region. In each step, it selects the region with the highest learning progress using the $\epsilon$-greedy \cite{sutton2018reinforcement} selection mechanism, and (object, action) pairs are sampled corresponding to the selected region for interaction. 

Learning progress (LP) of a region is calculated from the prediction performance change of the corresponding FM after a given learning update cycle: 

\begin{equation}\label{eq:LP}
    LP_n (t+1) = \gamma_n(t+1) - \gamma_n(t+1 - \theta),
\end{equation}
where  $\gamma_n$ indicates the mean error of FM\textsubscript{n} at update cycle $t$, and is calculated as follows:
\begin{equation}\label{eq:mean_error}
    \gamma_n(t+1) = \dfrac{\sum_{i=0}^{\theta}e_n(t+1-i)}{\theta+1}
\end{equation}
where the error of $n^{th}$ region $e_n(t)$ is calculated by the MSE between the predicted effect $E_{pred}$ and the observed effect $E_{obs}$. The window parameter $\theta$ allows the system to capture the trend of the errors by averaging them within a given learning period and prevents the fluctuations from affecting the IM signal. \changed{In general, a small $\theta$ makes the LP measure highly unstable, whereas a large $\theta$ makes the LP changes less precise \cite{ugur2016emergent}. Although in our experiments, $\theta$ is empirically set to 16 for computational convenience, it is possible to implement a mechanism to determine it (e.g., see\cite{schillaci2020tracking}) automatically.}  
  
\section{Experiment Setup}\label{sec:experimentsetup}
The experiment setup was simulated in CoppeliaSim \cite{rohmer2013v}, where a six-degrees-of-freedom robot arm with a gripper (UR10)\footnote{\url{https://www.universal-robots.com/products/ur10-robot/}} was chosen as the manipulator to be used in the experiments. \changed{In order to capture the basic interaction infants create with their environment, a simple setup with three types of objects was created, which the robot could interact with through its {\em reach and enclose} actions.} The details of the objects used, the action parameters, and the outcome \changed{representation are} given below.

\begin{figure}[t]
    \centering
    \includegraphics[width=\columnwidth]{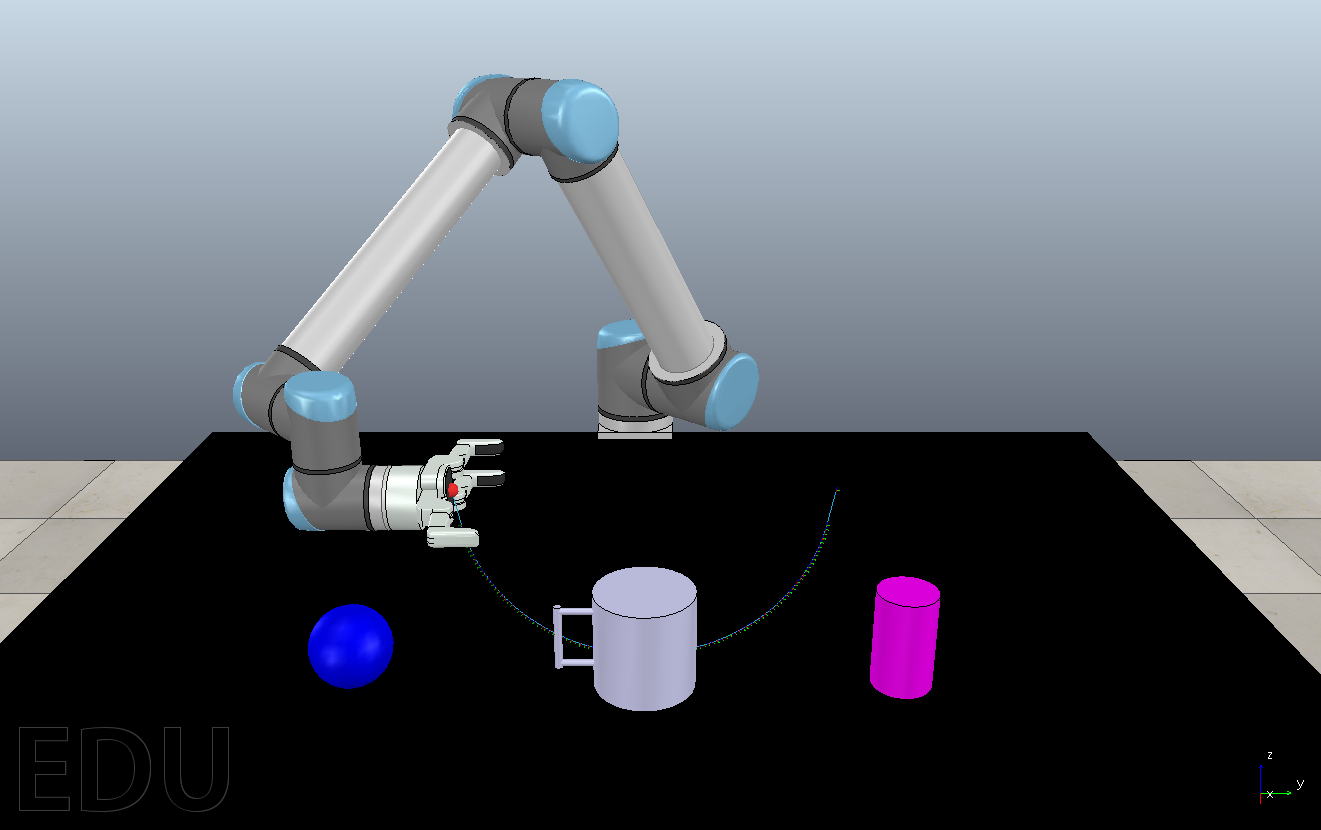}
    \caption{The experiment setup. A manipulator robot interacts with one of three types of objects in a table-top environment.}
    \label{fig:m_experiment}
\end{figure}

\paragraph*{Objects}
Three objects were used in the experiments with some changing sizes: a cup, a cylinder, and a sphere (Fig~\ref{fig:m_experiment}). The cup has a fixed height ($15$ cm) and radius ($7.5$ cm) and has a handle that is $12.5$ cm apart from the center with a length of $10$ cm. The orientation of the cup is changed around the vertical axis within $[0,2\pi]$, i.e., the position of the handle varies around the body of the cup. Cylinders have a fixed height of $h=15$ cm and radius within the range of $[1.5,7.5]$ cm, and spheres have a radius within the range of $[3,7.5]$ cm. A simulated Kinect camera is positioned on top of the table to record $128 \times 128$ top-down depth images of the objects.

\paragraph*{Actions}
{}
The end-effector of the robot follows a semi-circular trajectory that has start and end-points with the same elevation from the tabletop. The closest point of the trajectory to the table is the halfway point, and it has a fixed offset from the surface of the table to avoid collision between the end-effector and the table. The semi-circular trajectory is defined by the radius of the semi-circle $r_{path}=[26, 31]$ cm and a z-orientation within $\phi_{path} = [0,2\pi]$ radians. $\phi_{path}$ controls the approach direction of the end-effector to the object, i.e., determines the via point that the end-effector will pass while approaching the object (see Fig.~\ref{fig:m_actionparam}). When the end-effector interacts with the object, it takes one of three states: \textit{closed}, \textit{half-open}, and \textit{open}, and the fingers are enclosed, similar to a reflex, as soon as the object is contacted. In summary, the action parameter vector $A$ is composed of a 5-dimensional vector $(r_{path},\phi_{path},closed,half\-open,open)$  where the last three parameters are binary and represent the one-hot encoding of the gripper state.

\paragraph*{Outcome}
The outcome is defined as the change in the 3D position of the object, together with the (sine and cosine of the) orientation angle change with respect to the vertical axis: $O = (\Delta x,\Delta y, \Delta z,\sin{\Delta\phi_z}, \cos{\Delta\phi_z})$. The outcome is calculated by taking the difference between the first and final pose of the object. We used sine and cosine values of $\Delta\phi_z$ to ensure continuity at the fundamental boundaries of the domain of sine and cosine. Note that, even if the robot executes the action with \textit{open} and \textit{half-open} end-effector aperture configurations, it may not be able to grasp and raise the object. This can be caused by misalignment of the object size, object pose, end-effector pose, and simulation noise. For example, if the robot approaches with an open end-effector to the cup from the side of its handle, due to the contact of the handle with the robot fingers, the object rotates and is pushed out of the finger enclosure; hence it can not be grasped even if the fingers are enclosed.

For each interaction, the simulation scene is reset, and the parameters of the selected type of object and actions are sampled from their corresponding intervals. Overall, the dataset consists of three different object types with three different end-effector states, each consist of $5184$ interactions, in total $3\times3\times5184=46656$ interactions.

\begin{figure*}[t]
\begin{center}
\includegraphics[width=\textwidth]{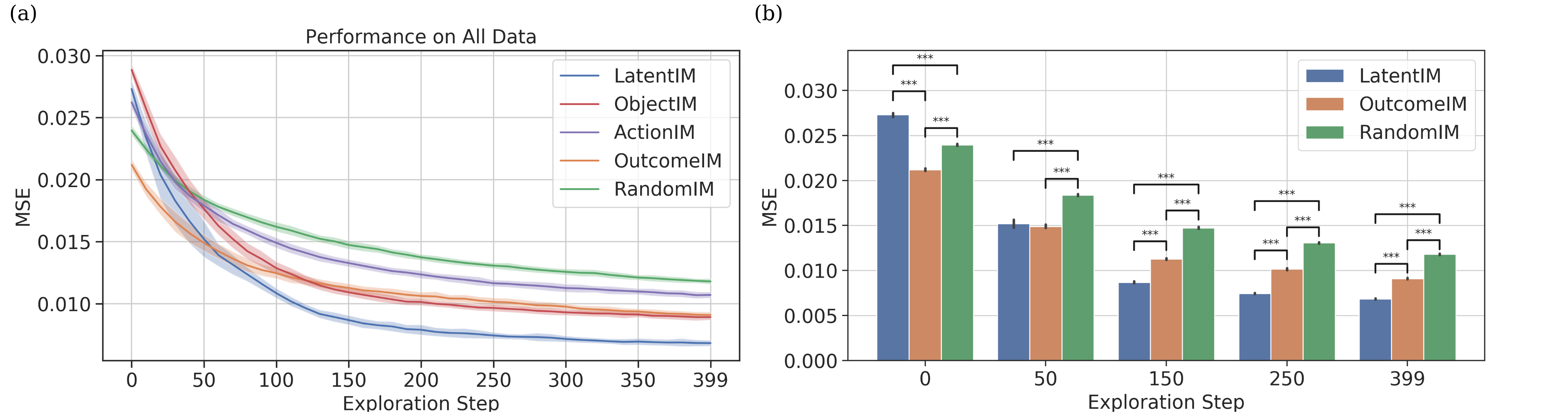}
\end{center}
\caption{ Comparison of the prediction performances of \latentmethod{}, \outcomemethod{}, \objectmethod{}, \actionmethod{} and \randommethod{}. (a) shows the change in the average MSE during the IM-based active exploration phase of 40 independent runs. The shaded areas show the standard deviation. (b) shows the same data but comparing only \latentmethod{}, \outcomemethod{}, and \randommethod{}, showing the statistical significance.}\label{fig:r_comp_combined}
\end{figure*}

\begin{figure}[b]
    \centering
    \includegraphics[width=\linewidth]{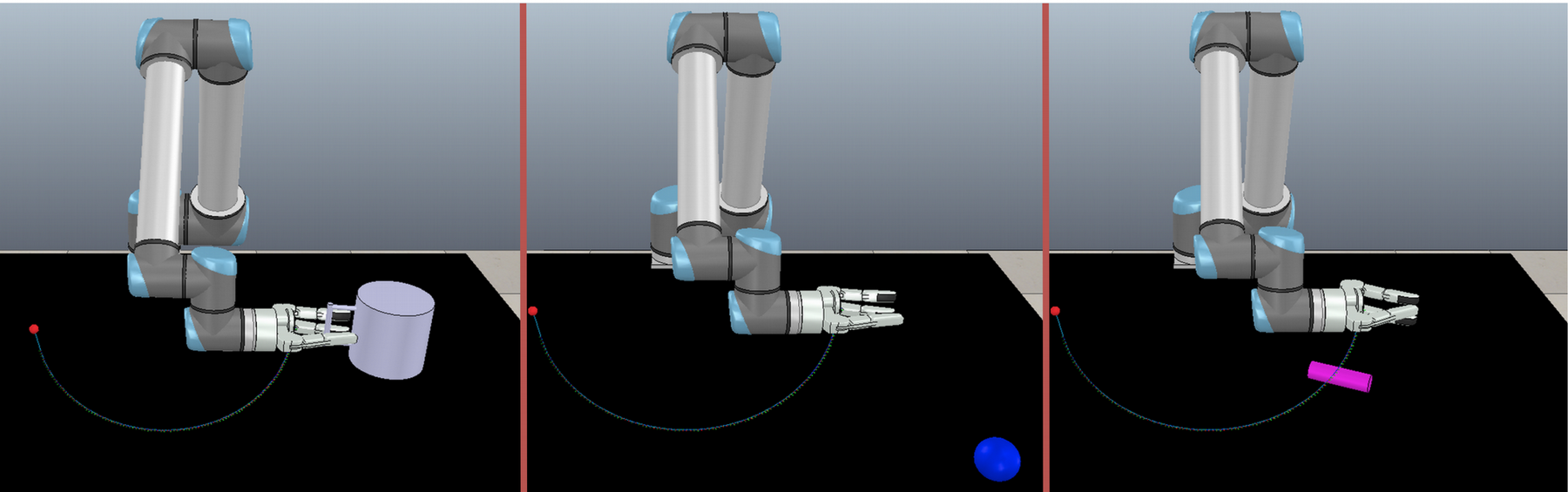}
    \caption{The final snapshots from sample interactions of $(l\_pinch)$, $(n\_pinch)$, and $(n\_close)$ regions.}
    \label{fig:r_execution}
\end{figure}{} 

\paragraph*{System Hyper-Parameters}
\begin{itemize}
    \item The convolutional auto-encoder, whose bottleneck layer (8D) serves as the object features ($I_{enc}$), consists of stacks of convolutional layers followed by batch normalization and max-pooling operations, with channel numbers 512, 256, 128, 64, 32, 16, and 8. It is trained using binary cross-entropy as the reconstruction loss and Adadelta \cite{zeiler2012adadelta} optimizer.
    \item The initial bootstrapping phase uses 700 random interactions for region formation. After the bootstrapping phase, the FMs are initialized with an initial set of 128 interactions, and the selected ones by IM are continued to be trained with the past interactions plus newly sampled $\kappa=16$ interactions for $400$ exploration steps. The number of regions is set to $5$. Thus, the IM-Based active exploration phase uses approximately $7000$ data.
    \item The other parameters are set as follows: $\epsilon=0.3$, $\theta=16$. 
\end{itemize}

\section{Results}\label{sec:results}
In this section, we analyzed the results of our latent space based IM approach (\latentmethod{}), and compared it with the alternatives that use only object (\objectmethod{}), action (\actionmethod{}), and outcome (\outcomemethod{}) spaces in region partitioning with the same number of regions. \changed{These variants are the adapted versions of existing IM methods that utilize only object \cite{ugur2007curiosity}, action \cite{ugur2016emergent}, and outcome \cite{baranes2013active}.} As a basic baseline, we also provided the results of \randommethod{} that assigns regions to the data points randomly. We conducted experiments to answer the following questions:
\begin{enumerate}
    \item How does the method of region formation affect the overall performance? (Section \ref{ssec:r_overall})
    \item What is the \changed{exploration} order of prediction capabilities? (Section \ref{ssec:r_developmental})
    \item What is the effect of the different hyper-parameters (number of clusters and $\epsilon$), \changed{using inverse models, and using different implementations of the components,} on overall performance? (\appref{apen})
\end{enumerate}{}

\subsection{Regions formed by \latentmethod{}}\label{ssec:r_skills}

We analyzed the regions formed in the latent space and identified the following segregation: Region 1 includes actions with \textit{half-open} gripper and objects with no change in $z$ position; region 2 includes actions with \textit{open} gripper and objects with no change in $z$ position; region 3 includes \textit{half-open} gripper and objects with changes in $z$ position; region 4 includes \textit{open} gripper and objects with changes in $z$ position;  region 5 includes \textit{closed}. Considering these characteristics, we named region 1 as non-lifting pinch-grasp $(n\_pinch)$, region 2 as non-lifting power-grasp $(n\_power)$, region 3 as lifting pinch-grasp $(l\_pinch)$, region 4 as lifting power-grasp $(l\_power)$ and region \changed{5} as non-lifting $(n\_close)$. Note that these labels are given to help the reader, and the system did not use any given labels. Sample snapshots from interactions of $(l\_pinch)$, $(n\_pinch)$, and $(n\_close)$ regions are provided in Fig.~\ref{fig:r_execution}.

\subsection{Comparison of Overall Performances}\label{ssec:r_overall}

To investigate the effect of the region formation on the overall performance of the system, we analyzed five different models, namely \latentmethod{}, \objectmethod{}, \actionmethod{}, \outcomemethod{}, and \randommethod{}. Fig.~\ref{fig:r_comp_combined}(a) shows the average MSE calculated with Eq.(\ref{eq:weightedMSE}) over 40 independent runs:
\changed{
    \begin{equation}\label{eq:weightedMSE}
        MSE = \dfrac{\sum_{i=1}^{N} n_i .  \psi_i}{\sum_{i=1}^N n_i}
    \end{equation}
    \begin{equation}
        \psi_i = \dfrac{1}{n_i}\sum_{j=1}^{n_i}(E_{obs}^i - E_{pred}^i)^2,
    \end{equation}
    where  $\psi_i$ stands for the mean squared error made by region $i$; $N$ and $n_i$ indicate the number of regions and the number of data points in region $i$, respectively; $E_{obs}^i$ and $E_{pred}^i$ represent the observed and predicted outcomes in region $i$, respectively.}
    
Note that the initial training of FMs with 128 samples is not included in the plot. As presented in the figure, \latentmethod{} gives the lowest error among all five models. Following \latentmethod{}, \outcomemethod{}, and \objectmethod{} perform similarly; the only difference between those two is that \outcomemethod{} is better at the beginning, but \objectmethod{} shows a more rapid decrease in the MSE. It seems that among all the methods, \outcomemethod{} benefits from the initial set of interactions most, considering that it groups similar outcomes, i.e., the data distribution among its regions is more coherent than the others. 

Depending on the actions applied, similar objects may give rise to observe different outcomes, and similar outcomes may be observed by applying similar actions on different objects. Therefore, observing that the performance of \objectmethod{} is close to \outcomemethod{} while \actionmethod{} achieves a lower performance is an interesting result for us. This situation might be linked with using object-related information from the encoded representation, i.e., different actions might be more informative than the different $I_{enc}$ to determine the outcome. \objectmethod{} and \actionmethod{} are not included in the rest of the paper for the readability and clarity of the figures. We considered \outcomemethod{} as the competitor of our method and \randommethod{} as the baseline.

In Fig.~\ref{fig:r_comp_combined}(b), we present the statistical analysis of the differences between \latentmethod{}, \outcomemethod{}, and \randommethod{} taken from the different exploration steps. We ran an analysis of variance (ANOVA) to check whether the MSE distributions of these three approaches are different. Then we carried out post-hoc ANOVA tests, i.e., Tukey's HSD and Games-Howell Test, depending on the equal and non-equal variance cases, respectively. We found that after $t=50$, the performance of \latentmethod{} and \outcomemethod{} differs significantly $(p<0.001)$, \latentmethod{} giving more accurate predictions.

\subsection{\changed{Exploration} Order of Skill Prediction}\label{ssec:r_developmental}
\begin{figure}[ht!]
\begin{center}
\includegraphics[width=\linewidth]{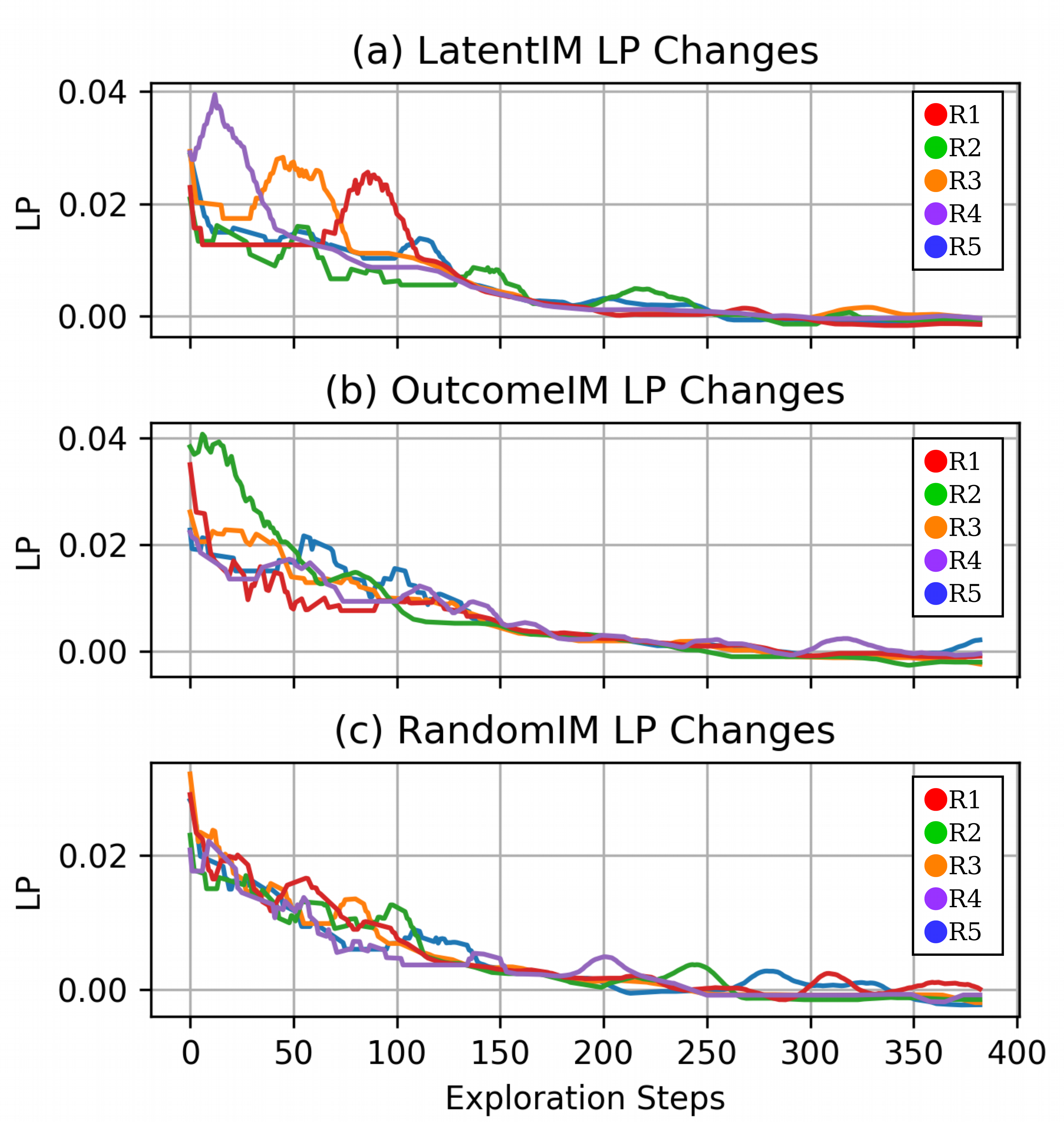}
\end{center}
\caption{Changes in learning progress of $5$ regions during the IM-based active exploration phase of a single run. The plot shows learning progress values for (a) \latentmethod{}, (b) \outcomemethod{}, and (c) \randommethod{}. Note that the learning progress signal is smoothed by a window ($\alpha=16$) to make it easier to view. For \latentmethod{} in (a), R1, R2, R3, R4 and R5 correspond $n\_pinch$, $n\_power$, $l\_pinch$, $l\_power$, and $n\_close$, respectively. For \outcomemethod{} and \randommethod{}, no clear correspondence can be observed.}
\label{fig:dev_singlerun}
\end{figure}

\begin{figure}[hb!]
\includegraphics[width=\columnwidth]{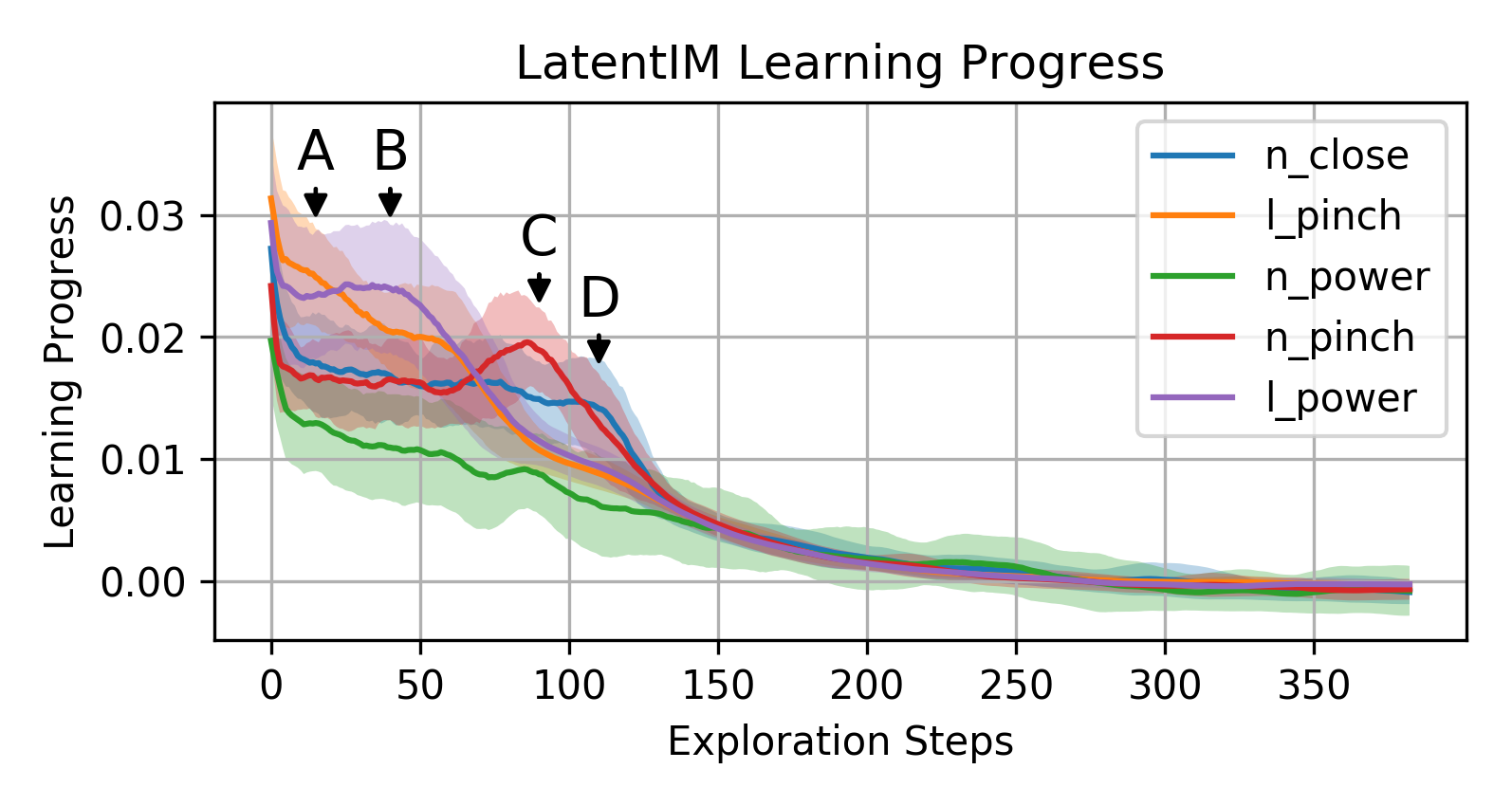}
\vspace{-0.8cm}
\caption{Change of mean learning progress of $5$ regions during the IM-based active exploration phase. The plot is smoothed by a time window ($\alpha=16$). The plot shows the mean learning progress values of the \latentmethod{} collected from $40$ experiments. The shaded areas show the standard deviation. Please refer to the text for the statistical analysis of the learning progress values at points A, B, C, and D.}
\label{fig:dev_multirun}
\end{figure}

In this subsection, we analyzed the \changed{exploration} order of regions and skill prediction that is regulated by the IM module. The analysis of the \changed{exploration} order with single runs of \latentmethod{}, \outcomemethod{} and \randommethod{} are presented in Fig.~\ref{fig:dev_singlerun}, and the average of 40 runs of \latentmethod{} is presented in Fig.~\ref{fig:dev_multirun}.

A positive LP value means that the predictions of the FM are improving. At each time step, the region with a higher LP value is most likely (due to the $\epsilon$-greedy region selection) to be selected for the exploration. Fig.~\ref{fig:dev_singlerun}(a) shows the learning progress values throughout the IM based active exploration phase of \latentmethod{} regions in a single run. The regions of \latentmethod{} show a clear ordering. It first explores the lifting grasps ($l\_power$ \& $l\_pinch$), then shifts its attention to $n\_pinch$, $n\_close$, and $n\_power$, respectively. Note that the order of skills may change across different runs due to the randomness involved in $\epsilon$-greedy region selection and sampling inside the regions. Detailed investigation and statistical analysis of the \changed{exploration} order formed by \latentmethod{} will be discussed later in this subsection.

In Fig.~\ref{fig:dev_singlerun}(b), \outcomemethod{} gives priority to $R2$, which corresponds to interactions with the cup object with $x,y$ position and orientation change. Following this, no clear ordering over the regions is observed. Similarly, \randommethod{} does not produce a distinctive \changed{exploration} order (Fig.~\ref{fig:dev_singlerun}(c)). Note that through the end of the exploration phase, for all the strategies, the LP values reach zero because of the strong FM predictors that can nevertheless learn their regions from the data points.

Fig.~\ref{fig:dev_multirun} shows the mean change in the learning progress of the regions for \latentmethod{} from 40 independent runs. The lines and shades correspond to the mean and standard deviations of the learning progress at the corresponding time steps. As shown in Fig.~\ref{fig:dev_multirun}, we observe a consistent ordering as $l\_pinch$, $l\_power$, $n\_pinch$, $n\_close$, and $n\_power$ on average. This ordering is reasonable because, when grasped, the orientation change is $\approx0$. However, when pushed, the object may turn around or tumble. Thus, when the robot lifts the object, the effect is more predictable than the rest; hence the corresponding region was easier to learn.

To evaluate the significance of the ordering presented in Fig.~\ref{fig:dev_multirun}, A Kruskal-Wallis test was performed on the learning progress of the five different regions. The differences between the learning progress distributions of the regions taken from the interval $t=[0, 139]$ are significant with $H(4)$, $p<0.01$. Following that, we also performed the Mann-Whitney U test to determine the significance of the learning progress values for the pairs of regions. In Fig.~\ref{fig:dev_multirun}, first $l\_pinch$ has the maximum learning progress value. Taken from that interval at $Point\:A$ the LP of $l\_pinch$ is significantly greater than of $l\_power$, $p<0.05$, and the LP of $l\_power$ is significantly greater than the rest with $p<0.001$. At the same time-step, $n\_pinch$ is not significantly different from $n\_close$, while the LP of $n\_close$ being significantly greater than of $n\_power$, $p<0.01$. Following the plot, we see the dominance of $l\_power$ over $l\_pinch$. Taken from that time window, at $Point\:B$, LP of $l\_power$ is significantly greater than of $l\_pinch$, $p<0.05$. While the LP of $l\_pinch$ is significantly greater than of $n\_close$ and  $n\_pinch$ with $p<0.001$, the difference between $n\_close$ and  $n\_pinch$ is not significant, both being greater than $n\_power$ with $p<0.01$. After the decrease of $l\_pinch$ and $l\_power$, a significant increase in $n\_pinch$ is visible in Fig.~\ref{fig:dev_multirun}. Being within that time window, at $Point\:C$, the LP of $n\_pinch$ is significantly greater than of $n\_close$, $p<0.05$. At the same time, LP of $l\_pinch$ is significantly less than of $n\_close$ with $p<0.05$ it is significantly greater than of $n\_power$, $p<0.001$. And finally, there is a short primacy of $n\_close$ over the rest, $Point\:D$, the LP of $n\_close$ is significantly greater than $n\_pinch$, $p<0.05$.

\section{Discussion and Conclusion}\label{sec:discussion}

\changed{Our experiment results show that the system can produce a sensorimotor learning curriculum that resembles some features of infants' sensorimotor development. This capability is attributable to the main ingredients of our work: (1) the latent space formed from object-action-outcome and (2) learning progress prioritization of local learning within the latent space.} Besides exhibiting developmentally plausible learning, the proposed system facilitates the development of better prediction ability by smartly distributing the exploration among the local learning modules defined over the blended latent space.

\textit{\changed{Exploration} order of prediction skills}. The proposed system developed prediction ability \changed{in a staged manner} for basic grasp actions before the prediction skill for purposeful push actions. This was an emergent feature realized through the coupling of learning progress based intrinsic motivation with the object-action-outcome blended space \changed{and resembles the order of emergence between grasp and push actions in infants}. However, the order within the grasp action types \changed{was not the same as infant development}. Since the precision pinch grasp requires finer control and precise movements in infants, it emerges later compared to the power grasp \cite{scharf2016developmental, cangelosi2015developmental}. Thus its related prediction ability should develop later as well. However, in our simulations, this order was reversed. The reason for this is easy to see, as in our simulations, the execution of the precision and power grasps has no differential difficulty due to the action parameterization used. Moreover, the robotic gripper used does not favor a power grasp, unlike a human hand that naturally conforms to the shape of the object once a basic hand enclosure is initiated \cite{twitchell1970}. On the contrary, the robotic gripper is more suitable for a precision pinch by design. Thus, in our experiments, the learning progress for pinch grasp turned out to be higher than that of power grasp, prioritizing the development of the prediction ability for precision pinch grasp. \changed{Yet, overall, the observed step-by-step improvement of skills in our experiments can be seen akin to the staged development observed during infancy \cite{scharf2016developmental}.}

\textit{Functional region emergence}. Another important feature the proposed system developed is that the regions formed over the blended space corresponded to well\changed{-}defined semantically meaningful action-outcome primitives. In particular, by analyzing the discovered regions,  we could identify  \textit{push}, \textit{grasp}, and \textit{near-grasp} actions. It could be questioned why clear object-based regions were not formed. The answer lies in the compact blended representation that finds the categorical actions (qualitatively different outcome yielding motor parameters) as a better descriptor for the triplets (action, object, outcome). This is comparable to the sensorimotor brain organization of primates for action, where the dorsal where/how pathway represents the objects in terms of features related to manipulation affordances, but not object \changed{identification}\cite{goodale_etal2005,goodale_milner1992}.

\textit{Superiority of the \changed{LatentIM over the other variants}}. 
\changed{Initially, it was not clear whether using a latent space to define the local models and exploring this latent space with IM} would yield better predictors. Yet, our experiments \changed{with forward model learning} showed that the latent space based IM significantly outperformed other IM approaches that use spaces only of the object, action, or outcome in terms of the learning speed and prediction accuracy.  \changed{Moreover, our experiments with inverse model learning (see \appref{apen}) also supported the superiority of \latentmethod{} over the other variants. Particularly, we have found that \latentmethod{} and \actionmethod{} outperformed the other three approaches, while \latentmethod{} showed a lower prediction error than \actionmethod{}.} Furthermore, the emergence of regions and \changed{exploration} order that we discussed above has only been observed clearly with the latent space based IM.

\textit{\changed{Implementation of the components.}} We believe that the general framework proposed well addresses the use of a diverse set of features observed during interactions in guiding IM exploration, whereas the particular implementation details might vary. For example, Variational Autoencoder (VAE), Gaussian Mixture Model (GMM), and Feed-Forward Neural Networks (FFNN) were used for latent space formation, the formation of exploration regions, and the effect prediction, respectively. \changed{Regarding these particular methods, we conducted experiments that used alternative methods with different capabilities (see \appref{apen}). Our analysis showed that dimensionality reduction, only with linear transformation (e.g., PCA) or clustering without variance information (e.g., K-Means) did not change the prediction performance significantly. Furthermore, the regions of \latentmethod{} formed with such methods appeared to be similar to those presented in Section~\ref{ssec:r_skills}. However, we found that using an FFNN with a linear activation function degraded the prediction performance for all approaches.}

\textit{\changed{Limitations.}} Finally, we would like to identify a number of limitations and possible future directions. First of all, the agents would encounter different situations and experience different interactions in a life-long learning scenario; therefore, mechanisms that allow assimilation and accommodation \cite{piaget1969psychology} of regions should be investigated. Regions found in our study reflect  object-action or action-outcome synergies; however, regions corresponding to individual objects, actions, or outcomes might also emerge in increasingly more complex environments. \changed{As stated in Section \ref{sec:experimentsetup}, our experiment setup is a simplified version of infants' play environments, e.g., the variety of the objects, actions, and outcomes are limited, and the robot fixates only one object at each interaction. In order to handle multiple objects, attention mechanisms would be required. In future work, along with an attention mechanism, the scalability of our method to more complex environments will be tested.} \changed{In addition, an experimental design that uses depth images for the outcome features is an exciting direction for our future research, especially in real-world scenarios with occlusions. As in \cite{laversanne2018curiosity, colas2019curious}, using the vision system for outcome features in a setup that includes distractor objects would be a worthwhile challenge for the intrinsic motivation studies. We would like to study this issue in future work with a real-world application.} Another point is that we used VAE for only latent space formation; however, the computational effort used in forming the latent space could have been exploited for also FM formation. In the current implementation, to ease the analysis, we decoupled latent space formation and FM learning by having separate mechanisms. As a future study, it would be interesting to explore the developmental progression of the system when a single neural architecture is used for both latent space formation and FM learning.

\section*{Acknowledgment}
This research was partially supported by JST CREST ``Cognitive Mirroring" [grant number: JPMJCR16E2] and TÜBİTAK (Scientific and Technological Research Council of Turkey) 2210-A scholarship. The authors would like to thank Serkan Bugur and Mert Imre for their comments on this study and this manuscript. We thank Mete Tuluhan Akbulut for proofreading this manuscript.

\ifCLASSOPTIONcaptionsoff
  \newpage
\fi



%
\bibliography{references}
\bibliographystyle{unsrt}

\appendix\label{apen}
In this section, we provide additional experiment results to examine the effect of the hyperparameters \changed{($\epsilon$, and the number of clusters), using an inverse model for the IM-based active exploration and using different implementations of the subsystems on the prediction performance.}

\begin{table}[h!]
\caption{ Mean MSE values for \latentmethod{}, \outcomemethod{} and \randommethod{} with different $\epsilon$ values.}\label{table:r_epsilon}
\begin{adjustbox}{width=\linewidth,center}
\begin{tabular}{l|l||l|l|l|}
\cline{2-5}
                                    & \textbf{F-Score} & \textbf{$\mu$(\latentmethod{})} & \textbf{$\mu$(\outcomemethod{})} & \textbf{$\mu$(\randommethod{})} \\ \hline
\multicolumn{1}{|l|}{$\epsilon$ = 0.0} &$ 1680.05          $&$ 0.007121              $&$ 0.009238              $&$ 0.011895$              \\ \hline
\multicolumn{1}{|l|}{$\epsilon$ = 0.5} &$ 3635.97          $&$ 0.006753              $&$ 0.009065              $&$ 0.011748$              \\ \hline
\multicolumn{1}{|l|}{$\epsilon$ = 0.7} &$ 3873.38          $&$ 0.006724              $&$ 0.009093              $&$ 0.011675$              \\ \hline
\multicolumn{1}{|l|}{$\epsilon$ = 1.0} &$ 4291.49          $&$ 0.006819              $&$ 0.008934              $&$ 0.011766$              \\ \hline
\end{tabular}
\end{adjustbox}
\end{table}

\begin{table*}[h!]
\caption{Mean MSE values for LatentIM, EffectIM and RandomIM with different number of clusters.}
\begin{adjustbox}{width=\textwidth,center}
\begin{tabular}{|c|c|c|c|c|c|c|c|c|c|}
\hline
\multirow{2}{*}{\begin{tabular}[c]{@{}c@{}}\# of \\ Clusters\end{tabular}} & \multicolumn{2}{c|}{\latentmethod{}} & \multicolumn{2}{c|}{\outcomemethod{}} & \multicolumn{2}{c|}{\randommethod{}} & \multirow{2}{*}{$\mu(L)-\mu(E)$} & \multirow{2}{*}{$\mu(L)-\mu(R)$} & \multirow{2}{*}{$\mu(E)-\mu(R)$} \\ \cline{2-7}
 & $\mu$ & $\sigma$ & $\mu$ & $\sigma$ & $\mu$ & $\sigma$ &  &  &  \\ \hline
2 & 0.009876 & 0.000293 & \textbf{0.009090} & 0.000232 & 0.010308 & 0.000302  &$ 0.00079^{***}$&$ -0.00043^{***}$&$ -0.00122^{***}$\\ \hline
3 & 0.009714 & 0.000354 & \textbf{0.008502} & 0.000231 & 0.011000 & 0.000210  &$ 0.00121^{***}$&$ -0.00129^{***}$&$ -0.00250^{***}$\\ \hline
4 & 0.008451 & 0.000352 & 0.008391 & 0.000203 & 0.011424 & 0.000258 		  &$ 0.00006^{ns}$&$ -0.00297^{***}$&$ -0.00303^{***}$\\ \hline
5 & \textbf{0.006836} & 0.000194 & 0.009195 & 0.000255 & 0.011787 & 0.000199  &$ -0.00236^{***} $&$ -0.00495^{***} $&$ -0.00259^{***}$ \\ \hline
6 & \textbf{0.008892} & 0.000260 & 0.010928 & 0.000264 & 0.012116 & 0.000186  &$ -0.00204^{***} $&$ -0.00322^{***} $&$ -0.00119^{***}$ \\ \hline
7 & \textbf{0.007666} & 0.000291 & 0.010956 & 0.000231 & 0.012588 & 0.000186  &$ -0.00329^{***} $&$ -0.00492^{***} $&$ -0.00163^{***}$ \\ \hline
\end{tabular}
\end{adjustbox}
\label{table:r_cluster}
\end{table*}

\subsection*{Effect of $\epsilon$ Parameter}
As explained in \ref{ssec:m_IML}, the $\epsilon$ parameter controls the ratio of exploration steps with random exploration to the ones with active exploration. We conducted $N=30$ experiments for each $\epsilon$ value and verified our results with One-Way ANOVA $F(2,87)$ tests followed by Tukey's HSD post-hoc on pairs (\latentmethod{} vs. \outcomemethod{}, \latentmethod{} vs. \randommethod{}, and \outcomemethod{} vs. \randommethod{}). The test yields that each pair is different with $p < 0.001$. 

In Table~\ref{table:r_epsilon}, we present the prediction errors of \latentmethod{}, \outcomemethod{}, and \randommethod{} with different values of the $\epsilon$ parameter. For all the $\epsilon$ values, we observe that the \latentmethod{} gives the lowest error among the other two.  We also observe that except for $\epsilon=0$, the performance of \latentmethod{} does not change significantly. 

\begin{figure}[t!]
    \centering
    \includegraphics[width=\linewidth]{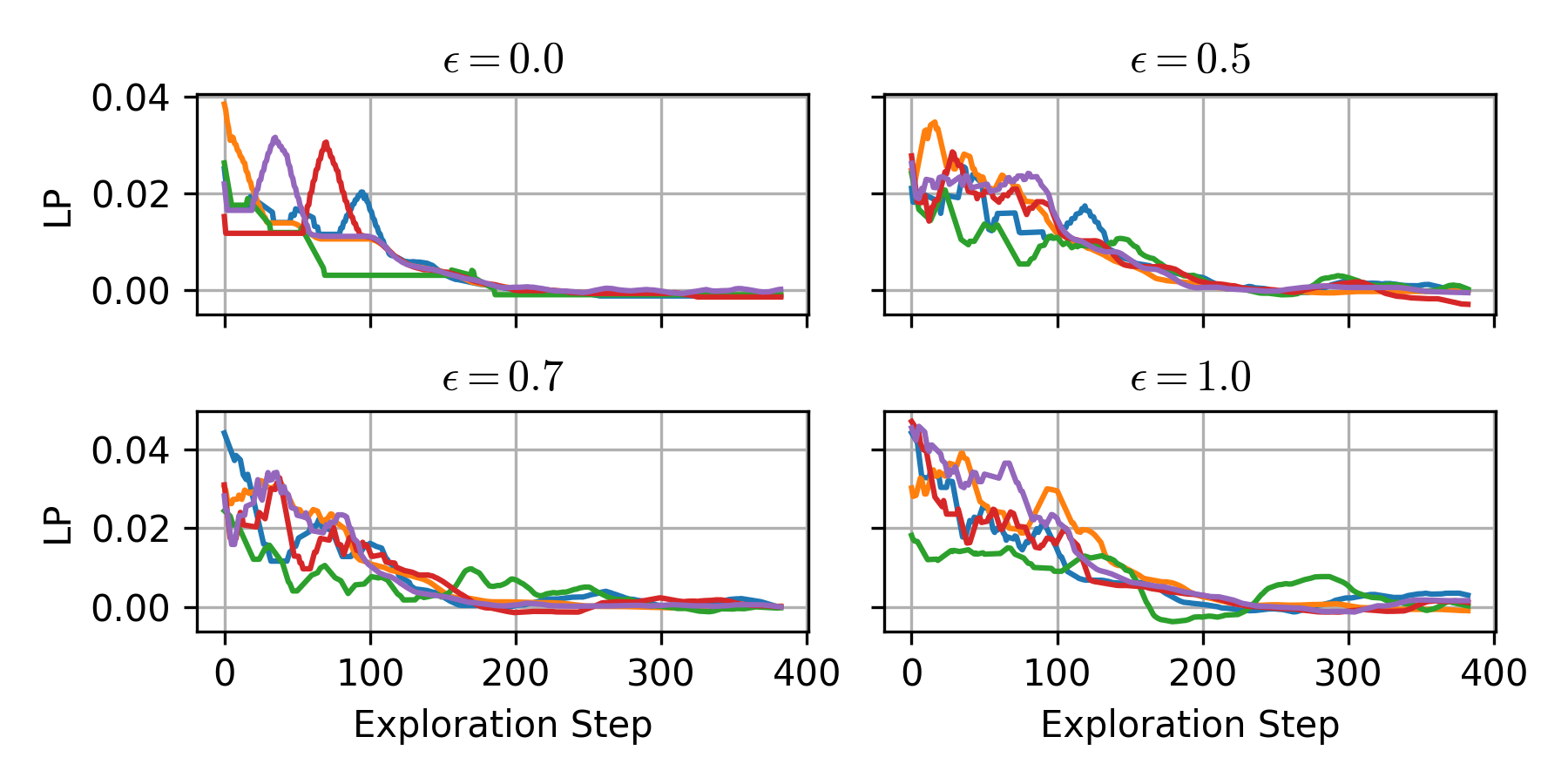}
    \caption{Learning progress plots of the regions of \latentmethod{} with different $\epsilon$ parameters. The plots are smoothed with time windows ($\alpha=16$).}
    \label{fig:r_epsilon_lp}
\end{figure}

In Fig.~\ref{fig:r_epsilon_lp}, we present the single run learning progress changes of \latentmethod{} with different $\epsilon$ conditions. Increasing values of $\epsilon$ prevents seeing an ordering between different skills and does not provide a benefit for the predictor performance.

{}

\subsection*{Effect of Number of Clusters}
In Table \ref{table:r_cluster}, we present the performance comparisons of \latentmethod{}, \outcomemethod{}, and \randommethod{} with different numbers of clusters. For this experiment, we used the hyper-parameters given in Section~\ref{sec:experimentsetup}, except for the number of clusters, which was the independent variable. The results show the mean and standard deviations of the methods' MSE calculated by conducting $30$ independent experiments each. To check the statistical significance of our findings, we used the One-Way ANOVA test, followed by Tukey's HSD post-hoc analysis. We observe that with $5, 6,$ and $7$ clusters, \latentmethod{} gives a lower prediction error that is statistically significant with $p<0.001$.

\begin{figure}[b!]
    \centering
    \includegraphics[width=\linewidth]{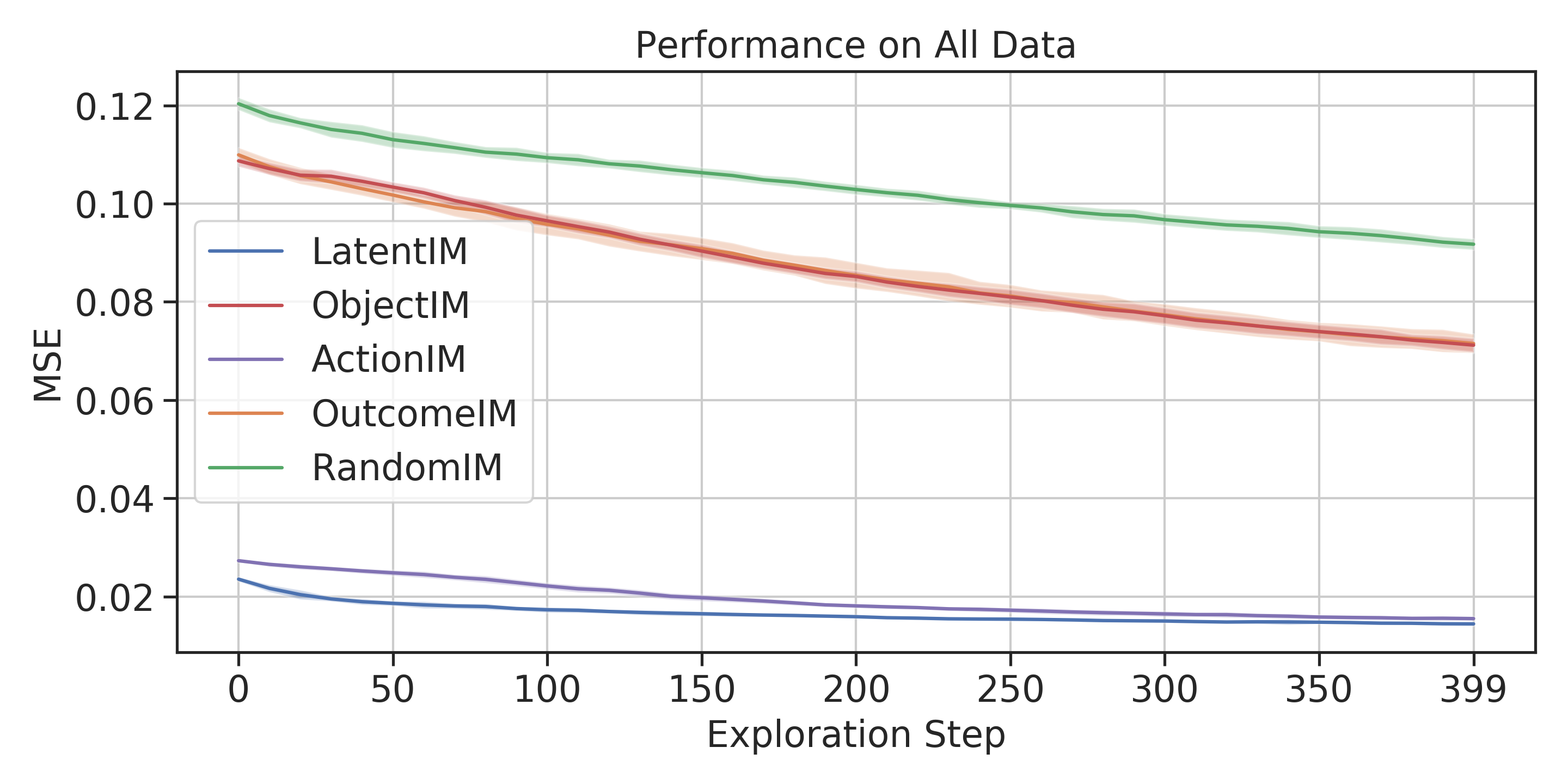}
    \caption{\changed{Comparison of the inverse models' prediction performances of \latentmethod{}, \outcomemethod{}, \objectmethod{}, \actionmethod{}, and \randommethod{}. The plot shows the change in the average MSE during exploration from 30 independent runs. The shaded areas show the standard deviation.}}
    \label{fig:app_inverse}
\end{figure}{}

\changed{\subsection*{Inverse Model Learning}
We also examined the performance of our method in inverse model learning where an inverse model $([I_{enc}, E] \mapsto A)$ was used instead of a forward model $([I_{enc}, A] \mapsto E)$. In order to analyze the performance of our intrinsic motivation strategies, we used the prediction error of the inverse model to measure the learning progress for the IM-based active exploration.}

\changed{Fig.~\ref{fig:app_inverse} shows the average weighted MSE of the 30 repeated runs for each model. We see that the performances of \latentmethod{} and \actionmethod{} are significantly better than the other three methods, where \latentmethod{} $(\mu=0.014418)$ and \actionmethod{} $(\mu= 0.015498)$ are almost tied with each other. The performances of \outcomemethod{} and \objectmethod{} are worse because the regions that they form do not differentiate the end-effector configurations \textit{(open, half-open, and close)}. In contrast, the regions of \latentmethod{} and \actionmethod{} are discriminated by the end-effector configurations, i.e., each region consisting of only one end-effector configuration. The differentiation based on end-effector configurations is crucial for the inverse model performance; for example, using the close end-effector configuration for lifting the object can not produce the desired outcome. In conclusion, \latentmethod{} performed better than the other variants on both forward and inverse model learning.}

\changed{\subsection*{Different Implementations of the Components}
To evaluate the effectiveness of the model components, we analyzed prediction errors and \latentmethod{} regions with different implementations of the components than those used in the proposed method. We conducted controlled experiments where the model components were replaced with alternative ones that have different capabilities. Table~\ref{table:subsytems} presents the prediction errors of each approach with the given combination of implementations. The first row shows the prediction errors (40 experiments) of the proposed method presented in the main text. The underlined components in the following rows indicate the replacements. Each experiment consisted of 10 independent runs.}

\begin{table}[b!]
\caption{MSE and standard deviation values for \latentmethod{},\outcomemethod{}, \randommethod{}, \objectmethod{}, \actionmethod{} with different implementations of the components.}
\begin{adjustbox}{width=\linewidth,center}
\begin{tabular}{|c|c|c|l|c|c|}
\hline
\begin{tabular}[c]{@{}c@{}}Dim.\\ Reduc.\end{tabular} & Clustering & \begin{tabular}[c]{@{}c@{}}Forward\\ Model\end{tabular} & \multicolumn{1}{c|}{Approach} & MSE & STD \\ \hline
\multirow{5}{*}{VAE} & \multirow{5}{*}{GMM} & \multirow{5}{*}{\begin{tabular}[c]{@{}c@{}}NN\\ ReLU\\ 512 Units\end{tabular}} & \textbf{LatentIM} & \textbf{0.0068} & 0.000237 \\ \cline{4-6} 
 &  &  & OutcomeIM & 0.0091 & 0.000234 \\ \cline{4-6} 
 &  &  & RandomIM & 0.0118 & 0.000222 \\ \cline{4-6} 
 &  &  & ObjectIM & 0.0089 & 0.000230 \\ \cline{4-6} 
 &  &  & ActionIM & 0.0107 & 0.000248 \\ \hline
\multirow{5}{*}{\underline{PCA}} & \multirow{5}{*}{GMM} & \multirow{5}{*}{\begin{tabular}[c]{@{}c@{}}NN\\ ReLU\\ 512 Units\end{tabular}} & \textbf{LatentIM} & \textbf{0.0083} & 0.000321 \\ \cline{4-6} 
 &  &  & OutcomeIM & 0.0091 & 0.000197 \\ \cline{4-6} 
 &  &  & RandomIM & 0.0117 & 0.000125 \\ \cline{4-6} 
 &  &  & ObjectIM & 0.0090 & 0.000245 \\ \cline{4-6} 
 &  &  & ActionIM & 0.0108 & 0.000201 \\ \hline
\multirow{5}{*}{\underline{AE}} & \multirow{5}{*}{GMM} & \multirow{5}{*}{\begin{tabular}[c]{@{}c@{}}NN\\ ReLU\\ 512 Units\end{tabular}} & \textbf{LatentIM} & \textbf{0.0078} & 0.000212 \\ \cline{4-6} 
 &  &  & OutcomeIM & 0.0090 & 0.000249 \\ \cline{4-6} 
 &  &  & RandomIM & 0.0119 & 0.000184 \\ \cline{4-6} 
 &  &  & ObjectIM & 0.0089 & 0.000272 \\ \cline{4-6} 
 &  &  & ActionIM & 0.0108 & 0.000151 \\ \hline
\multirow{5}{*}{VAE} & \multirow{5}{*}{GMM} & \multirow{5}{*}{\begin{tabular}[c]{@{}c@{}}NN\\ \underline{Linear}\\ 512 Units\end{tabular}} & LatentIM & 0.0284 & 0.000225 \\ \cline{4-6} 
 &  &  & OutcomeIM & 0.0213 & 0.000114 \\ \cline{4-6} 
 &  &  & RandomIM & 0.0232 & 0.000245 \\ \cline{4-6} 
 &  &  & ObjectIM & 0.0298 & 0.000178 \\ \cline{4-6} 
 &  &  & ActionIM & 0.0259 & 0.000201 \\ \hline
\multirow{5}{*}{VAE} & \multirow{5}{*}{\underline{K-Means}} & \multirow{5}{*}{\begin{tabular}[c]{@{}c@{}}NN\\ ReLU\\ 512 Units\end{tabular}} & \textbf{LatentIM} & \textbf{0.0066} & 0.000213 \\ \cline{4-6} 
 &  &  & OutcomeIM & 0.0094 & 0.000262 \\ \cline{4-6} 
 &  &  & RandomIM & 0.0119 & 0.000382 \\ \cline{4-6} 
 &  &  & ObjectIM & 0.0084 & 0.000326 \\ \cline{4-6} 
 &  &  & ActionIM & 0.0106 & 0.000235 \\ \hline
\end{tabular}
\end{adjustbox}
\label{table:subsytems}
\end{table}

\changed{For the PCA case (Table~\ref{table:subsytems}, $2^{nd}$ row), we used three principal components to reduce the dimensionality of the concatenation of the object, action, and outcome vector. For the Autoencoder ($3^{rd}$ row), again, we used a three dimensional latent space. The encoder had an 18 dimensional input layer followed by dense layers with 32, 16, 8, 4 hidden units and a tanh activation function. The decoder of the AE was symmetric to the encoder. It was observed that with Autoencoder and PCA methods, the regions of \latentmethod{} were similar to the regions presented in the main text. Thus, the prediction performance did not change significantly.}

\changed{For the forward models, we used linear activation function (Table~\ref{table:subsytems}, $4^{th}$ row) instead of ReLU. We observed that using linear activation made the MSE values for all the approaches higher, i.e., prediction performances of all the approaches became more inadequate compared to the ReLU case. Thus, we conclude that, with our experiment setup, the forward models need to approximate non-linear relationships.}

\changed{As a clustering method, K-Means (Table~\ref{table:subsytems}, $5^{th}$ row) was used instead of GMM. Following the proposed method, we set the number of clusters as five. The regions of \latentmethod{}, as well as the prediction errors, were quite similar to the GMM case. Furthermore, we also analyzed the regions of other approaches and found only slight changes in the percentages of the object types in each region. Thus, their performances also did not change significantly.}

\end{document}